\documentclass[11pt,a4paper]{article}
\usepackage[hyperref]{emnlp2020}
\usepackage{times}
\usepackage{latexsym}
\usepackage{amssymb}

\usepackage{microtype}

\aclfinalcopy 

\usepackage{graphicx}
\usepackage{color}
\usepackage{url}

\title{ETC: Encoding Long and Structured Inputs in Transformers}

\author{Joshua Ainslie, Santiago Onta\~{n}\'{o}n, Chris Alberti, Vaclav Cvicek, \\ {\bf Zachary Fisher, Philip Pham, Anirudh Ravula, Sumit Sanghai, Qifan Wang, Li Yang} \\
  Google Research \\
  \{jainslie, santiontanon, chrisalberti, vcvicek, \\ zachfisher, phillypham, braineater, sumitsanghai, wqfcr, lyliyang\}@google.com 
}

\date{}

\begin{document}\maketitle
\begin{abstract}
Transformer models have advanced the state of the art in many Natural Language Processing (NLP) tasks. In this paper, we present a new Transformer architecture, {\em Extended Transformer Construction} (ETC), that addresses two key challenges of standard Transformer architectures, namely scaling input length and encoding structured inputs. To scale attention to longer inputs, we introduce a novel global-local attention mechanism between global tokens and regular input tokens. We also show that combining global-local attention with relative position encodings and a {\em Contrastive Predictive Coding} (CPC) pre-training objective allows ETC to encode structured inputs. We achieve state-of-the-art results on four natural language datasets requiring long and/or structured inputs.
\end{abstract}

\section{Introduction}


Models based on Transformers~\cite{vaswani2017attention}, such as BERT~\cite{devlin2018bert}, or other variants~\cite{yang2019xlnet,lan2019albert,raffel2019exploring} have yielded state-of-the-art results in many NLP tasks such as language modeling~\cite{child2019generating,sukhbaatar2019adaptive,rae2019compressive,kitaev2020reformer}, question answering~\cite{lan2019albert,beltagy2020longformer}, and summarization~\cite{zhang2019hibert}. 
We present the {\em Extended Transformer Construction} (ETC) architecture\footnote{Source code and pre-trained checkpoints for ETC can be found at \url{http://goo.gle/research-etc-model}}, targeting two limitations of the original models: (1) scaling input length, (2) encoding structured inputs.

The computational and memory complexity of attention in the original Transformer scales quadratically with the input length, typically limiting input length to around 512 tokens. While 512 might be enough for some tasks (e.g., co-reference resolution seems to benefit from even smaller input lengths~\cite{joshi2019bert}), this is problematic in others. Consider question answering (QA) tasks that require reasoning across multiple documents (e.g., the {\em HotpotQA} dataset~\cite{yang2018hotpotqa}) all of which must simultaneously fit in the model input. Other examples are summarization, or QA on long documents. Many approaches have been proposed to address this, like hierarchical processing~\cite{zhang2019hibert}, sparse attention~\cite{child2019generating}, and segment-level recurrence~\cite{dai2019transformer}.

A second limitation is that few models focus on {\em structured inputs}, by which we refer to any underlying graph or hierarchical structure among the input tokens. Although ETC can encode more general graph structure, in this work we focus on representing hierarchical structure in NLP tasks, not usually modeled by Transformer models. For example, text is organized into sentences and paragraphs, and while these have a sequential order, different input documents might not hold any order between them (e.g., the HotpotQA dataset). Additionally, web text contains markup and is laid out using a DOM tree, giving additional structure. We show ETC can represent these and other types of structure, like linking different entity mentions.



To address these challenges, we present a novel attention mechanism called {\em global-local attention}, which divides the input into two sequences (which we call the {\em global} input and the {\em long} input). This mechanism introduces local sparsity to reduce the quadratic scaling of the attention mechanism. When this is coupled with relative position encodings~\cite{shaw2018self}, it allows for handling structured inputs in a natural way. 
Additionally, unlike previous Transformer extensions, ETC can be initialized from existing pre-trained standard BERT models (which together with a GPU/TPU-friendly implementation, allows for efficient model training)\footnote{An exception to this is {\em Longformer}~\cite{beltagy2020longformer}, a new model developed concurrently to ETC, which also allows initialization from BERT/RoBERTa.}.  Our results show that initializing from RoBERTa~\cite{liu2019roberta} significantly improves performance. Finally, we show that by adding a pre-training {\em Contrastive Predictive Coding} (CPC) task~\cite{oord2018representation}, performance improves even further for tasks where structure is important, as CPC plays the role of a masked language model (MLM) task, but at a sentence level of granularity.

We report experiments on four datasets: {\em Natural Questions} (NQ)~\cite{kwiatkowski2019natural},  HotpotQA~\cite{yang2018hotpotqa}, WikiHop~\cite{welbl2018constructing}, and OpenKP (part of MS MARCO)~\cite{xiong2019open}, which have long and/or structured inputs. We set a new state of the art in all of them.

Moreover, although in this paper we strictly focus on ETC, in a related model called BigBird~\cite{zaheer2020big}, we experimented with an alternative set of ideas to handle long inputs and its extensions to a decoder for text generation. The focus of BigBird is on the idea of adding random sparse attention patterns to global-local attention, and on showing under which conditions models like BigBird/ETC are universal approximators of sequence functions and are Turing complete. While the key ideas and techniques required to achieve the state-of-the-art results mentioned above for QA tasks are the focus of this paper, the reader is referred to the BigBird work for a joint evaluation of ETC (referred to as BigBird-ETC in that work) and the idea of random sparse attention patterns.


\section{Background}\label{sec:background}

\begin{figure*}[t!]
	\includegraphics[width=0.965\textwidth]{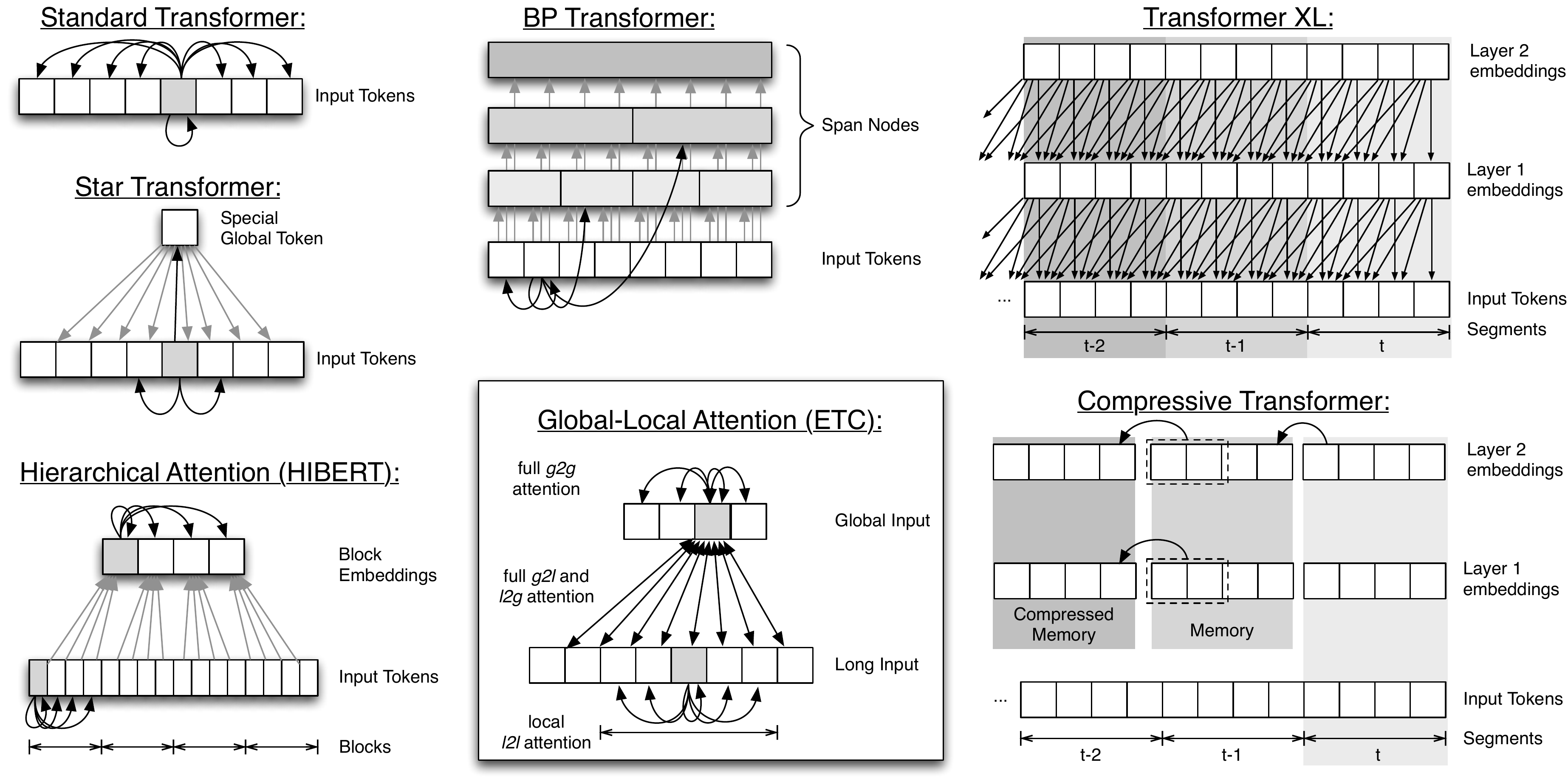}
	\centering
	\caption{An illustration of mechanisms to scale attention to long inputs, including our proposed model, ETC.}
	\label{fig:attention-mechanisms}
\end{figure*}



Many variants of the original Transformer model~\cite{vaswani2017attention} have been proposed for scaling up training~\citep[RoBERTa,][]{liu2019roberta}, the internal representation~\citep[ALBERT,][]{lan2019albert}, or both~\cite[T5,][]{raffel2019exploring}, outperforming BERT~\cite{devlin2018bert} in tasks such as GLUE~\cite{wang2018glue}, SQuAD~\cite{rajpurkar2016squad} or RACE~\cite{lai2017race}. However, these models typically limit inputs to $n=512$ tokens due to the $O(n^2)$ cost of attention. 
We classify prior approaches to scale up attention into four categories: sparse attention, recurrence, hierarchical mechanisms, and compressed attention.

{\bf Sparse Attention} involves limiting each token to attend to a subset of the other tokens. For example, the {\em Sparse Transformer}~\cite{child2019generating} used predefined attention patterns for both text and image generation. 
They showed that attending only to previous pixels in the same row or column was enough to generate high quality images, while keeping attention cost at $O(n\sqrt{n})$. 
In the {\em Adaptive Attention Span Transformer}~\cite{sukhbaatar2019adaptive} each attention head is associated with a decaying learnable masking function, which limits the number of tokens it can attend to. 
They show that lower layers learn to use short attention spans, and only in higher layers are attention spans longer. Sparse attention has also been used to increase the interpretability of attention heads by allowing attention to assign exactly zero weight to certain input tokens~\cite{correia2019adaptively}. The {\em Reformer}~\cite{kitaev2020reformer} model 
finds the nearest neighbors of the attention query (those input tokens that would result in the highest attention weights) using locality sensing hashing~\cite{andoni2015practical} and only uses those for attention. This reduces attention cost to $O(n \,  \mathit{log}(n))$. The {\em Routing Transformer}~\cite{roy2020efficient} learns dynamic sparse attention patterns using online $k$-means, reducing complexity to $O(n^{1.5})$. Finally, the most related approach to the work presented in this paper is {\em Longformer}~\cite{beltagy2020longformer}, developed concurrently to ETC, and which features a very similar global-local attention mechanism as ETC's but does not directly encode graph or hierarchical structure (more detailed comparison in Section~\ref{sec:etc}).

{\bf Recurrence} incorporates elements of recurrent neural networks into Transformer models to lengthen their attention span. {\em Transformer-XL}~\cite{dai2019transformer} takes this approach, dividing the input sequence into segments and then processing these segments one at a time. At each layer, the model attends to the layer immediately below for both the current and previous input segments. The effect is that layer $k$ is influenced by the current segment and the $k-1$ previous segments, as shown in the top-right of Figure~\ref{fig:attention-mechanisms}.

In {\bf Hierarchical Mechanisms} the input sequence is split into {\em blocks} that are ingested independently to produce summary embeddings that represent the whole block. Then, separate layers ingest the concatenation of these embeddings. For example, HIBERT~\cite{zhang2019hibert} uses this idea at the sentence level for extractive summarization (illustrated in the bottom-left of Figure~\ref{fig:attention-mechanisms}). Hierarchical attention in Transformers has also been applied to other NLP tasks such as neural machine translation~\cite{maruf2019selective}. 
Moreover, notice that this idea of processing the input hierarchically is not specific to Transformer models, and it has been applied to recurrent neural network models both at the level of sentences~\cite{yang2016hierarchical,miculicich2018document} and blocks~\cite{shen2018bi}. 
%

{\bf Compressed Attention} takes the idea of hierarchical attention one step further by selectively compressing certain parts of the input. The {\em BP-Transformer}~\cite{ye2019bp} model builds a binary partitioning tree over the input, and only lets the model attend to the leaves (the raw tokens) for nearby tokens, and higher nodes in the tree (summaries of groups of tokens) as tokens grow more distant (see Figure~\ref{fig:attention-mechanisms}, middle top). Other ideas include {\em memory compressed attention}~\cite{liu2018generating} where groups of $k$ tokens are compressed via a convolution filter before they are attended to, and the {\em Star Transformer}~\cite{guo2019star}, where each token can attend only to its immediate left/right neighbors and to a separate special auxiliary token that represents a summary of the whole input (see Figure~\ref{fig:attention-mechanisms}, left).
The {\em Compressive Transformer}~\cite{rae2019compressive} integrates this idea into Transformer-XL by compressing tokens in the input that are far away. The model benefits from detailed attention to nearby tokens, while using summarized information for more distant tokens (see Figure \ref{fig:attention-mechanisms}, lower right). 

\section{Extended Transformer Construction}\label{sec:etc}

Our model follows the original Transformer architecture~\cite{vaswani2017attention}, with key modifications to tackle long and structured inputs: relative position encoding, global-local attention, and a CPC pre-training task, explained below. In this paper, we consider only the encoder side of the Transformer, and leave the decoder for future work.

\subsection{Relative Position Encoding}\label{sec:relativepe}

Inspired by the work of \citet{shaw2018self}, ETC replaces absolute position encodings with relative position encodings, 
which provide information about the relative position of tokens in the input sequence with respect to one another. Given the input sequence $x = (x_1, ..., x_n)$, we can see it as a labeled fully connected and directed graph, where $l_{ij}$ is the label of the edge that connects $x_i$ to $x_j$. 
Given a maximum clipping distance $k$, \citeauthor{shaw2018self} define $2k+1$ {\em relative position labels}: $l_{-k}, ..., l_{k}$. The label of the edge between two input tokens depends only on their relative position $j-i$. For input pairs with $j-i \geq k$, label $l_k$ is given, and with $j-i \leq -k$, $l_{-k}$ is given. Each label then becomes a learnable vector $a_l^K$, which modifies the attention mechanism (equations in the next section)\footnote{In the work of Shaw et al., a second $a_l^V$ vector was used, but their ablations showed it may not affect performance.}.

\begin{figure*}[t!]
	\includegraphics[width=0.9\textwidth]{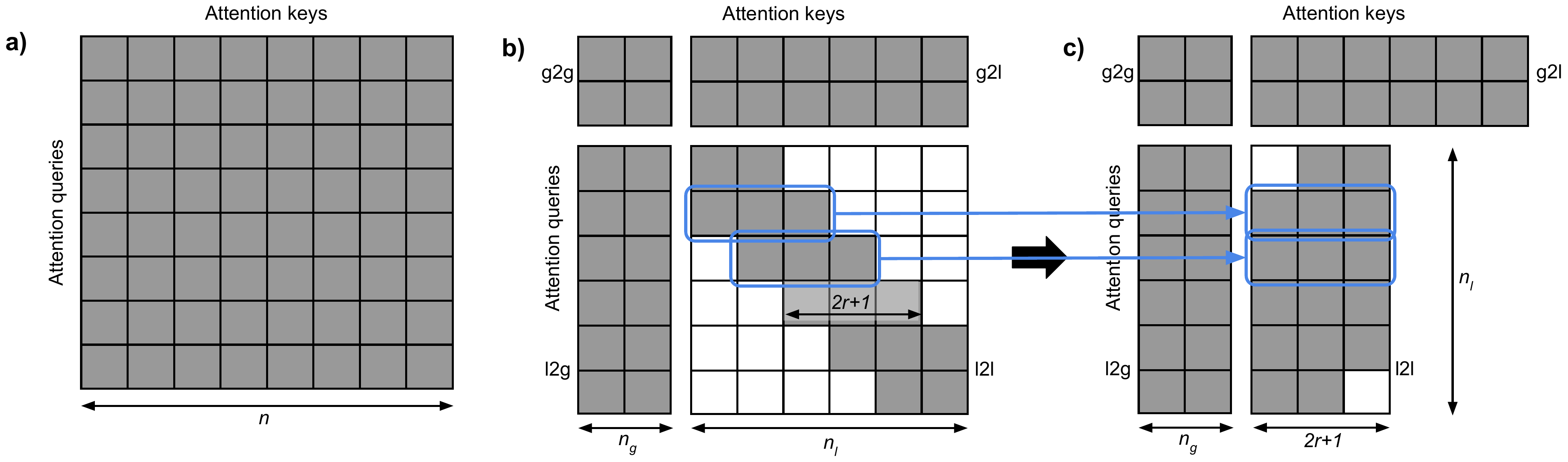}
	\centering
	\caption{Sparsity diagram showing which attention queries (rows) can attend to which attention keys (columns) a) for standard Transformer attention with input size $n$; b) for global-local attention with input sizes $n_g$, $n_l$, and radius $r$; c) how the l2l attention piece is reshaped into a much smaller attention matrix, limited by local radius.}
	\label{fig:etc-attention}
\end{figure*}

Relative position encodings are independent of input length, so it is easy to adapt a model to greater input lengths than seen during pre-training. As other recent work~\cite{shaw2019generating},  ETC's attention mechanism uses relative position labels not just for relative positions in a sequence but also to express arbitrary pairwise token relations useful for structured inputs, as explained below.

\subsection{Global-Local Attention}

Global-local attention is a generalization of several of the models presented above. 
%
ETC receives two separate input sequences: the {\em global input} $x^g = (x^g_1, ..., x^g_{n_g})$ and the {\em long input} $x^l = (x^l_1, ..., x^l_{n_l})$. Typically, the long input contains the input a standard Transformer would receive, while the global input contains a much smaller number of auxiliary tokens ($n_g \ll n_l$). Attention is then split into four separate pieces: {\em global-to-global} (g2g),  {\em global-to-long} (g2l),  {\em long-to-global} (l2g), and  {\em long-to-long} (l2l). 
Attention in the l2l piece (the most computationally expensive piece) is restricted to a fixed radius $r \ll n_l$. To compensate for this limited attention span, the tokens in the global input have unrestricted attention, and thus long input tokens can transfer information to each other through global input tokens. Accordingly, g2g, g2l, and l2g pieces of attention are unrestricted. 

This concept is illustrated in Figure~\ref{fig:etc-attention}, where each cell (row $i$, column $j$) is shaded grey if token $x_i$ can attend to token $x_j$. As we can see, in a regular Transformer, attention is unrestricted (full $n \times n$ attention). ETC, illustrated in Figure~\ref{fig:etc-attention}b, however, restricts the l2l piece to a local radius, significantly reducing computational and memory complexity for very long inputs. Conceptually, the l2l attention piece is reshaped into a $n_l \times (2r+1)$ matrix as illustrated in Figure~\ref{fig:etc-attention}c.\footnote{In practice, for GPU/TPU efficiency, a different reshaping occurs that yields identical outputs (see the appendices).}

If $r = 1$ and $n_g = 1$, we recover exactly the Star Transformer (Section~\ref{sec:background}). Similarly, placing all the tokens in the global input and setting $n_l = 0$ yields standard Transformer attention. Attention in ETC is $O(n_g (n_g + n_l) + n_l (n_g + 2r+1 ))$. If we assume $n_g = O(2r+1)$, we see attention is linear in the size of the long input: $O(n_g^2 + n_g n_l)$.

To provide flexible attention and help with structured inputs, per-instance Boolean attention matrices $M^{g2g}$, $M^{g2l}$, $M^{l2g}$, and $M^{l2l}$ exist, with zeroes for those pairs of tokens that should not attend to one another. 
Each g2g attention head works as follows. Given the global input $x^g = (x^g_1, ..., x^g_{n_g})$, which is a sequence of token representations $x^g_i \in \mathbb{R}^{d_x}$, the output of attention is $z^g = (z^g_1, ..., z^g_{n_g})$, where each $z^g_i \in \mathbb{R}^{d_z}$ is calculated as follows:
\[z^g_i = \sum_{j=1}^{n_g} \alpha^{g2g}_{ij} x^g_j W^V\]
\[\alpha^{g2g}_{ij} = \frac{\exp(e^{g2g}_{ij})}{\sum_{\ell=1}^n \exp(e^{g2g}_{i\ell})}\]
\[e^{g2g}_{ij} = \frac{x^g_i W^Q (x^g_j W^K + a_{ij}^K)^T}{\sqrt{d_z}} - (1 - M^{g2g}_{ij})C\]
\noindent where: $M^{g2g}$ is a binary attention mask, $W^Q$, $W^K$, and $W^V$ are learnable weight matrices, and $a_{ij}^K$ are learnable vectors representing the relative position labels, and $C$ is a large constant ($C = 10000$ in our experiments to follow the same convention as BERT). 
Attention for the other 3 pieces is analogous. We experiment with having separate $W^K$ and $W^V$ across all four attention pieces, or sharing them. And for $W^Q$, we experiment with having one for g2g and g2l, and a separate one for l2g and l2l; or sharing them also. To recover BERT as a special case when $r$ is large enough to remove sparsity, attention is actually only split into 2 pieces internally instead of 4, as g2g+g2l can be computed jointly (top half of Figure \ref{fig:etc-attention}c), and l2g+l2l can also be computed jointly (bottom half of Figure \ref{fig:etc-attention}c). A single softmax is used to jointly calculate $\alpha^{g2g}_{ij}$ and $\alpha^{g2l}_{ij}$, and another for $\alpha^{l2g}_{ij}$ and $\alpha^{l2l}_{ij}$.

Thus, the output of global-local attention is a sequence of length $n_g$ and one of length $n_l$. These sequences go through a layer normalization and feed forward layer in the same way as in the standard transformer. 


\subsection{Long Inputs and Global-Local Attention}\label{subsec:longinputs}

\begin{figure}[t!]
	\includegraphics[width=\columnwidth]{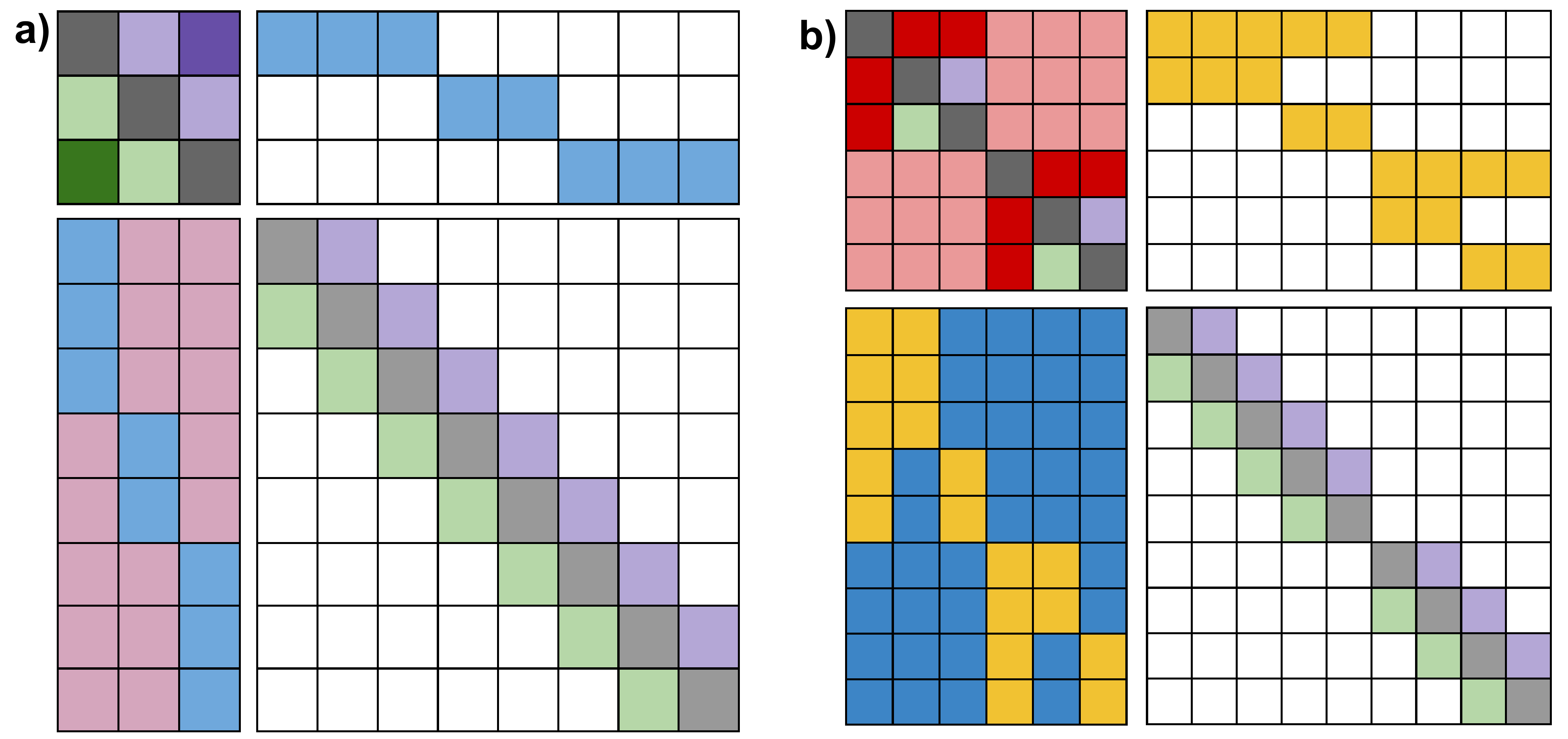}
	\centering
	\caption{Example attention patterns for handling (a) long inputs and (b) structured inputs. White background means attention is masked via $M$, and the other colors indicate different relative position labels.}
	\label{fig:etc-attention2}
\end{figure}

Let us illustrate how ETC can be used to encode long inputs. 
A general way to handle long inputs in ETC is to place the entire sequence of input tokens (e.g., word pieces) in the long input, and then assuming some sort of division into segments (e.g., sentences), place one auxiliary token in the global input per segment in the long input. We then use one relative position label to link the global segment tokens with the word piece tokens that belong to them, and a different label for those that do not. Moreover, as we will show in the experiments below, we have seen that using the $M^{g2l}$ attention masks to perform hard masking in one direction (g2l) can bring performance gains in some datasets. This last asymmetric hard-masking is illustrated in Figure \ref{fig:etc-attention2}a, where we used different colors to indicate different relative position labels. In this way, although tokens in the long input can only attend to the local neighborhood defined by the radius $k$, they can indirectly attend to all other tokens in the input sequence via the global tokens. 
%

\subsection{Structured Inputs}\label{subsec:structure}

A standard Transformer resembles a graph neural network~\cite{scarselli2008graph} over a fully connected graph $g$; see~\citet{ye2019bp}. Thanks to the combination of global-local attention and relative position labels, ETC exploits this relation to encode {\em structured inputs}. 
Given the input $x = (x_1, ..., x_n)$, we use the term {\em structure} to refer to the relations that exist between the tokens in $x$. When $x$ is a plain ordered sequence, the only relation is the {\em sequential order} of tokens, which is the only structure captured by BERT (encoded by absolute position encodings, used to modify attention). We define {\em structured inputs} as those that have additional relations between the tokens beyond sequential order. In principle, we could think of inputs with arbitrary graph structure (such as chemical molecule graphs), but here we focus on structure in NLP tasks.

ETC is particularly well suited to capture hierarchical structure thanks to three mechanisms. First, as originally conceived, the vocabulary of relative position labels is used to represent token relative positions. However, seeing a Transformer as a graph neural network over a graph $g$ (with one vertex per token in $x$, and edges representing their relations), we can expand this vocabulary to label some edges with labels for relations such as {\em is-a}, {\em part-of}, or others. Second, the division between long and global input induces a natural structure where the global input contains summary tokens of sets of tokens in $x$ (a 2-level hierarchy). However, we can also have tokens summarizing sets of summary tokens (constructing a 3-level hierarchy, or beyond). Third, if some pairs of tokens should not have an edge between them, this can be captured with the $M^{g2g}$, $M^{g2l}$, $M^{l2g}$, $M^{l2l}$ masks. An illustration of all these concepts is shown in Figure~\ref{fig:etc-attention2}b, which uses masking and relative position labels to represent a context-sentence-token hierarchy that includes within-context order of sentences but no order between contexts. 
Another example would be social community graphs structure, where we could partition the graph into components, use $M^{l2l}$ to constrain attention to within components, and add per-component global tokens, linked to allow information to propagate from one component to another in a hierarchical way.



\subsection{Pre-training Tasks}

We use two pre-training tasks: (1) a masked language model (MLM) task with {\em whole word masking} (if one word piece token is masked, then all other tokens of the same word are masked); and (2) instead of using BERT's next sentence prediction (NSP), we adapt Contrastive Predictive Coding (CPC)~\cite{oord2018representation} for ETC. 

The goal of CPC is to predict subsequent inputs in latent space, i.e., to predict internal hidden representations of blocks of tokens. We adapted this idea in ETC by using global input sentence summary tokens. Given an input sequence containing $n$ sentences, we mask all the tokens corresponding to a subset of sentences (but leave the sentence summary tokens in the global input). We then train the model to minimize the difference between the hidden representation of the global sentence summary tokens for the masked sentences with respect to that of a global summary token that can see the unmasked sentence and nothing else. We use a Noise Contrastive Estimation (NCE) loss as in the work of Oord et al. (\citeyear{oord2018representation}) (details in the appendices). 

Having described ETC, we can now compare it with Longformer~\cite{beltagy2020longformer}, which uses a similar attention mechanism, except Longformer has a single input sequence with some tokens marked as {\em global} (the only ones that use full attention). The key differences are that (1) ETC's combination of global-local attention with relative position encodings and flexible masking enables it to encode structured inputs in a similar way as graph neural networks do; (2) global tokens in Longformer are never pre-trained with anything like our CPC loss, and thus their use must be learned during fine-tuning. 

\subsection{Lifting Weights from Existing Models}

ETC and BERT share enough similarities that BERT parameters are useful to perform a warm start. The parameters are compatible because the global-local attention mechanism includes BERT as a special case if the input is small enough or the local radius is large enough to eliminate sparsity.
Moreover, when lifting weights from BERT into an ETC model with separate $W^Q$, $W^K$, and $W^V$ projection matrices, BERT's parameters are just copied over to the different matrices of ETC. 

Although pre-training is still required to adapt the weights to use global tokens and relative position encodings, we show that initializing from RoBERTa results in significant performance improvements compared to pre-training from scratch. Specifically, we initialized from the RoBERTa checkpoints reported in the work of Rothe et al.~\cite{rothe2020leveraging}.


\section{Empirical Evaluation}\label{sec:experiments}

This section focuses on evaluating our two main contributions: (1) long inputs, and (2) structure in text inputs, as well as initialization from existing BERT models. 
%
%
We chose four datasets (Table \ref{tbl:datasets}) 
with long inputs or interesting input structure.

{\bf NQ}~\cite{kwiatkowski2019natural}: in Google's {\em Natural Questions} (NQ) dataset the input consists of a question and a full Wikipedia article. The task is to identify both a {\em short answer} (a few words from the article) and a {\em long answer} (e.g., a whole paragraph), if they exist within the article (and otherwise, return null answers). Performance is measured based on the F1 score of the model predictions with respect to the human generated answers. 

{\bf HotpotQA}~\cite{yang2018hotpotqa} is a question answering dataset where the goal is to combine evidence from multiple contexts. We use the {\em distractor} setting, where 10 paragraphs are provided: two of them contain useful information to answer the question, and the rest are distractors. The task is both to answer the question, and also to identify the supporting facts that are relevant to answer the questions (at a sentence granularity). 

{\bf WikiHop}~\cite{welbl2018constructing} is similar in structure to HotpotQA. The contexts correspond to portions of Wikipedia articles, and the goal is to answer about properties of an entity that cannot be found in the entity's article. Each instance contains a query, a collection of candidate answers, and a collection of contexts from which to obtain information to select among the candidate answers.

{\bf OpenKP}~\cite{xiong2019open} is a keyphrase extraction dataset. Each document contains up to 3 short keyphrases to be identified. We selected this dataset as the input is not flat text sequences, but websites, including the hierarchical and spatial relations between the different DOM elements on the website, as well as other visual properties.

\begin{table}[tb]\centering 
\resizebox{\columnwidth}{!}{
\begin{tabular}{l|rr|rrrr} 
& \multicolumn{2}{c|}{{\em Instances}} & \multicolumn{3}{c}{{\em Instance length}} \\
{\em Dataset} & \multicolumn{1}{c}{\em Training} & \multicolumn{1}{c|}{\em Dev} & \multicolumn{1}{c}{\em Median} &
\multicolumn{1}{c}{\em 95\%} & \multicolumn{1}{c}{\em Max}  \\ \hline
NQ          & 307373 & 7830 & 4004 & 17137 & 156551 \\
HotpotQA    & 90447 & 7405 & 1227 & 1810 & 3560 \\
WikiHop     & 43738 & 5129 & 1541 & 3994 & 20337 \\
OpenKP      & 133724 & 6610 & 761 & 4546 & 89183 \\
\end{tabular}		
}
\caption{Dataset stats (length in word piece tokens).}
\label{tbl:datasets} 
\end{table}



\begin{table*}[tb]\centering 

\begin{small}
\begin{tabular}{lclr|cc} 
{\em Model} & {\em Input length} & {\em Configuration} & {\em \#Params} & {\em Long answer F1} & {\em Short answer F1}  \\ \hline
BERT-base    &   512     &  & 110M                   & 0.634     & 0.475 \\
BERT-large    &   512     & & 340M                     & 0.647     & 0.527 \\
RikiNet & 512       & lifting from RoBERTa$_{\mathit{large}}$ & - &  0.753 & {\bf 0.593} \\ 
\hline
ETC	    &   512	    & shared, no CPC, no hard g2l                   & 109M  & 0.645     & 0.478 \\
ETC	    &   4096	    & shared, no CPC, no hard g2l               & 109M      & 0.692	& 0.497 \\
ETC	    &   4096	    & fixed blocks, shared, no CPC, no hard g2l               & 109M      & 0.697	     & 0.508 \\
ETC	    &   4096	    & shared, no hard g2l                 & 109M    & 0.717	     & 0.524 \\
ETC	    &   4096	    & shared                & 109M     & 0.721		     & 0.514 \\
ETC	    &   4096    & -                & 166M     & 0.725		    & 0.522 \\ \hline
ETC	    &   8192    &                  & 166M    & 0.740	    & 0.542 \\
ETC	    &   4096	& 2x local radius    & 166M & 0.737		    & 0.530 \\ 
ETC	    &   4096	& 2x relative vocab   & 166M$^*$  & 0.733	    & 0.532 \\ 
ETC	    &   4096	& 2x pre-training     & 166M  & {\bf 0.746}	    & {\bf 0.558} \\ \hline
ETC-large	&   4096	&                & 539M  & 0.761	     & 0.565 \\
ETC-large   &   4096    & lifting from RoBERTa  & 558M & {\bf 0.782}	& 0.585 \\
\end{tabular}	
\end{small}
\caption{Empirical results on the dev sev set for the {\em Natural Questions} (NQ) dataset. Best results for {\em base} and {\em large} models highlighted. BERT-large results obtained from \citet{alberti2019bert}. $^*$ although not visible due to rounding to the closest million, doubling the relative position encoding vocabulary adds about 600k parameters.}
\label{tbl:results-nq} 
\end{table*}

\subsection{Training Configuration}

We use two basic configurations: {\em base} and {\em large}. Base uses 12 layers, 768 hidden size, 12 attention heads, local attention radius $r=84$, and relative position maximum distance $k=12$. Large uses 24 layers, 1024 hidden size, 16 heads, $r=169$, and $k=24$. We used 128, 230 and 460 global tokens for models with 512, 4096 and 8192 long input size respectively in NQ\footnote{With gradient checkpointing, ETC can scale beyond this, but we limit our experiments to 8192 tokens for this paper.}, 256 global tokens in HotpotQA, 430 in WikiHop, and 512 in OpenKP.

{\bf Pre-training:} We place all word piece tokens in the long input and add one auxiliary token per sentence to the global input.
We defaulted to BERT's 30k English uncased word piece vocabulary. Models were pre-trained using the original BERT datasets, except that documents with fewer than 7 sentences were filtered out. Unless stated otherwise, base models were pre-trained with the same total number of tokens as the original BERT, and for large models, twice as many. We used the {\em LAMB} optimizer~\cite{you2019large} with learning rate set to $\sqrt{8}\times10^{-3}$. 

{\bf Fine-tuning:} we put all input tokens in the long input (CLS, question, and context tokens for QA datasets), and use relative position labels to encode structure (see Section~\ref{subsec:structure}). Global input has a CLS token, tokens mirroring the question tokens in long, and one summary token per paragraph/sentence (or VDOM block in OpenKP). OpenKP had no CLS nor question tokens. For WikiHop, we also add one global token per candidate answer, and used a different relative position label to link these tokens to their string-matched mentions in the text (more details in the appendices).

\begin{table*}[tb]\centering 
\begin{small}
\begin{tabular}{lclr|cc} 
{\em Model} & {\em Input length} & {\em Configuration} & {\em \#Params} & {\em HotpotQA} & {\em WikiHop}\\ 
 &  &  & & Ans. F1 / Sup. F1 & Acc. \\ \hline
Longformer  & 4096  &  & 149M$^*$ & 0.743 / 0.844 &  75.0  \\ 
Longformer-large  & 4096  &  & 435M$^*$ &  0.788 / 0.860\footnotemark &  77.6  \\ \hline
ETC	        &   4096	& flat structure, no CPC, no hard g2l   & 166M & 0.722 / 0.857     & 70.0   \\
ETC	        &   4096	& flat structure                        & 166M & 0.748 / {\bf 0.870}     & 70.7     \\
ETC	        &   4096	& no CPC                                & 166M & 0.747 / 0.866     & 73.0    \\
ETC	        &   4096	& no hard g2l                           & 166M & 0.743 / 0.864     & {\bf 75.9}  \\
ETC	        &   4096	& shared                                & 109M & 0.733 / 0.866     & 73.7   \\
ETC	        &   4096    & -                                     & 166M & {\bf 0.751} / 0.869	  & 73.2     \\ \hline
ETC-large	&   4096    &                                      & 539M & 0.798 / 0.890	    & 77.0    \\
ETC-large	&   4096    & lifting from RoBERTa                  & 558M & {\bf 0.813} / {\bf 0.894}	    & {\bf 79.8}    \\
\end{tabular}		
\end{small}
\caption{Empirical results on HotpotQA and WikiHop (dev set results). $^*$Longformer parameter counts provided by the authors via personal communication.}
\label{tbl:results-hop} 
\end{table*}

\begin{table*}[tb]\centering 
\begin{small}
\begin{tabular}{lclr|c} 
{\em Model} & {\em Input length} & {\em Configuration} & {\em \#Params} & {\em OpenKP F1@3} \\ \hline
RoBERTa-JointKPE    & 512   &  & - & 0.398 \\ \hline
ETC	    &   512     & fixed blocks, no CPC, no hard g2l, no visual features      & 166M & 0.399 \\
ETC	    &   4096	& fixed blocks, no CPC, no hard g2l, no visual features      & 166M & 0.400 \\
ETC	    &   4096	& no CPC, no hard g2l, no visual features                         & 166M & 0.400 \\
ETC	    &   4096	& no hard g2l, no visual features  & 166M & 0.400 \\
ETC	    &   4096	& no visual features  & 166M & 0.402  \\
ETC	    &   4096    & -                         & 166M & 0.409 \\ 
ETC	    &   4096    & shared                         & 109M & 0.409 \\
ETC	    &   4096    & max loss                         & 166M &  {\bf 0.416} \\ \hline
ETC-large	&   4096    & max loss                    & 539M &  0.419 \\
ETC-large	&   4096    & max loss, lifting from RoBERTa                    & 558M & {\bf 0.423} \\
\end{tabular}		
\end{small}
\caption{Empirical results on OpenKP (dev set F1@3 results).}
\label{tbl:results-openkp} 
\end{table*}

\begin{table}[tb]\centering 
\begin{small}
\begin{tabular}{l|cc}
{\em Leaderboard} & Result & Position \\ \hline
NQ long answer  &  0.7778   & 1st \\
NQ short answer  &  0.5786   & 18th \\
HotpotQA Sup. F1  &   0.8909   & 1st \\
HotpotQA Overall  &   0.7362   & 3rd \\
WikiHop  &   0.8225   & 1st \\
OpenKP  &   0.4205   & 1st \\
\end{tabular}
\end{small}
\caption{Official leaderboard results for ETC at the time of submission.}
\label{tbl:results-leaderboard} 
\end{table}

\subsection{Results on the Dev Set}

{\bf NQ:} We used NQ to study the different parts of ETC via ablations. Results are shown in Table~\ref{tbl:results-nq}. The first three rows show baseline models: BERT-base, BERT-large, and RikiNet~\cite{liu2020rikinet} (one of the best models in the NQ leaderboard). 
BERT's performance is comparable to ETC using input length of 512. The smaller local radius of ETC (84) puts ETC at a disadvantage with respect to BERT, but other ETC improvements, such as dynamic whole word masking seem to compensate. 

The rest of Table~\ref{tbl:results-nq} shows performance under different ablations. Our default configuration (marked with a ``-'' in the configuration column) is ETC-base with long input length of 4096 tokens, using CPC, hard g2l masking, and separate $W^Q$, $W^K$, and $W^V$ matrices for long/global inputs. We tested the following ablations: {\em shared} (sharing all model parameters for attention across both the global and long inputs), {\em no CPC} (removing the CPC pre-training task), {\em no hard g2l} (not having a hard g2l mask), and {\em fixed blocks} (which configures the global input to just have one global token per 97 long input tokens, to keep the same proportion as without fixed-blocks, ignoring sentence boundaries, and not having any other tokens in the global input for pre-training or fine-tuning). 
Sharing $W^Q$, $W^K$, and $W^V$ and removing CPC significantly hurt the performance of ETC in NQ\footnote{Separate projection matrices were also found to be helpful in other models, like Longformer~\cite{beltagy2020longformer}.}. Using fixed blocks, surprisingly, seems to slightly help without CPC.

Increasing long input from 512 to 4096 significantly helped performance, and going to 8192 increased performance further to 0.740 / 0.542, highlighting the importance of longer inputs. Increasing the local radius, relative position vocabulary, or the amount of pre-training all helped performance (especially the latter, reaching 0.746 / 0.558). 
Moving to a large model also helped, especially when lifting from RoBERTa (both large models used the RoBERTa vocabulary). Lifting from RoBERTa achieved our best scores: 0.782 / 0.585, beating the best dev scores in the literature for long answer (compare with 0.754 / 0.593 for RikiNet). For short answer, we still lag behind RikiNet.




{\bf HotpotQA, WikiHop:} Table \ref{tbl:results-hop} shows our results in HotpotQA and WikiHop. We show two Longformer models as baselines (which is currently the state-of-the-art model in WikiHop), as well as ablations to study the effect of structure in the results. In particular, we consider a {\em flat structure} ablation where: (1) we do not break long input attention by context boundaries, (2) we limit relative position labels between global and long tokens to representing only sentence-level relationships (this removes any special attention in WikiHop between candidate answers and their mentions).

Our results show that both our base and large models outperform their corresponding Longformer models in both HotpotQA and WikiHop. Besides parameter counts, the main factors that can explain this difference in performance are the different pre-training strategies and the different handling of structure in ETC and Longformer.
Removing the CPC pre-training task, and not using a hard g2l mask significantly hurt the performance of the model in HotpotQA, going from a performance of 0.751 / 0.869 for the baseline model to 0.722 / 0.857 using none of those features. Using a flat structure (but keeping CPC and hard g2l) did not seem to hurt in HotpotQA.
WikiHop shows a slightly different picture, as it seems that hard g2l masking and especially flat structure hurt performance in this dataset. Our best model is the base configuration without hard g2l masking, which achieves an accuracy of 75.9. Interestingly, sharing $W^Q$, $W^K$, and $W^V$ seems to help performance in WikiHop. This is our smallest dataset, and maybe the added capacity of the model without sharing parameters leads it to overfit.

{\bf OpenKP:} Table \ref{tbl:results-openkp} shows our results on the OpenKP dataset, using RoBERTa-JointKPE~\cite{sun2020joint} as the baseline, which is currently \#1 in the leaderboard. This is an interesting structured dataset, and thus, we performed additional ablations to investigate the effect of removing such structural information. Our results show that even the most constrained ETC model already achieves very good performance (0.399), and scaling to 4096 length seems to give a slight boost. Using hard g2l also helps, and adding the visual features brings the largest benefit. Finally, we see that using a large model, and especially lifting weights from RoBERTa improve results significantly. As with WikiHop, sharing $W^Q$, $W^K$, and $W^V$ does not hurt performance. Our default model uses the first occurrence of a keyphrase, but we saw that using the maximum logit of all occurrences ({\em max loss}) improved results.

\subsection{Official Leaderboard Results}

Finally, Table~\ref{tbl:results-leaderboard} shows official results on the leaderboards of each dataset. The model submitted to the leaderboards was the model with best dev set results (shown at the bottom of the respective results tables, lifting weights from RoBERTa). We set a new state of the art in WikiHop and OpenKP, NQ long answer, and HotpotQA Support F1. Remarkably, our submissions were all single model, outperforming the leaderboard ensemble models.

\footnotetext{Better results were reported for Longformer-large using a 2 stage approach, reaching 81.0 / 85.8 \cite{beltagy2020longformer}, but our table shows single-model results only, for comparison.}

\section{Conclusions}\label{sec:conclusions}

This paper introduced the Extended Transformer Construction (ETC), an architecture designed to (1) scale up the input length (linearly with input), and (2) encode structured inputs. ETC allows lifting weights from existing BERT models, improving results significantly. The key ideas are a new global-local attention mechanism, coupled with relative position encodings and a CPC pre-training task.

We showed that significant gains can be obtained thanks to increased input sequence length. The ability to represent dataset structure in ETC further improves the model quality. We hypothesize that CPC helps the model train the usage of the higher-level global input summary tokens, as CPC plays a role akin to MLM, but at the global input level. Notice that although our datasets contain a limited amount of structure (compared to graph datasets), our experiments show that ETC was able to exploit this existing structure.  

As future work, we would like to 
investigate complementary attention mechanisms like those of Reformer~\cite{kitaev2020reformer} or Routing Transformer~\cite{roy2020efficient}, push scalability with ideas like those from RevNet~\cite{gomez2017reversible}, and study the performance of ETC in datasets with even richer structure. 

\section*{Acknowledgements} 
We would like to thank Anna Goldie, Bhargav Kanagal, Ilya Eckstein, Manan Shah, Nich Kwon, Vikram Rao Sudarshan, Joshua Maynez, Manzil Zaheer, Kelvin Guu, Tom Kwiatkowski, Kristina Toutanova, and D. Sivakumar for helpful discussions, support, comments, and feedback on earlier versions of this work.  We would also like to thank the Longformer authors (Iz Beltagy, Matthew E. Peters, Arman Cohan) for their useful feedback on earlier versions of this paper and for sharing parameter counts.

\bibliographystyle{acl_natbib}

\begin{thebibliography}{44}
\expandafter\ifx\csname natexlab\endcsname\relax\def\natexlab#1{#1}\fi

\bibitem[{Alberti et~al.(2019)Alberti, Lee, and Collins}]{alberti2019bert}
Chris Alberti, Kenton Lee, and Michael Collins. 2019.
\newblock A bert baseline for the natural questions.
\newblock \emph{arXiv preprint arXiv:1901.08634}.

\bibitem[{Andoni et~al.(2015)Andoni, Indyk, Laarhoven, Razenshteyn, and
  Schmidt}]{andoni2015practical}
Alexandr Andoni, Piotr Indyk, Thijs Laarhoven, Ilya Razenshteyn, and Ludwig
  Schmidt. 2015.
\newblock Practical and optimal lsh for angular distance.
\newblock In \emph{Advances in neural information processing systems}, pages
  1225--1233.

\bibitem[{Beltagy et~al.(2020)Beltagy, E.~Peters, and
  Cohan}]{beltagy2020longformer}
Iz~Beltagy, Matthew E.~Peters, and Arman Cohan. 2020.
\newblock Longformer: The long-document transformer.
\newblock \emph{arXiv preprint arXiv:2004.05150v1}.

\bibitem[{Child et~al.(2019)Child, Gray, Radford, and
  Sutskever}]{child2019generating}
Rewon Child, Scott Gray, Alec Radford, and Ilya Sutskever. 2019.
\newblock Generating long sequences with sparse transformers.
\newblock \emph{arXiv preprint arXiv:1904.10509}.

\bibitem[{Correia et~al.(2019)Correia, Niculae, and
  Martins}]{correia2019adaptively}
Gon{\c{c}}alo~M Correia, Vlad Niculae, and Andr{\'e}~FT Martins. 2019.
\newblock Adaptively sparse transformers.
\newblock \emph{arXiv preprint arXiv:1909.00015}.

\bibitem[{Dai et~al.(2019)Dai, Yang, Yang, Carbonell, Le, and
  Salakhutdinov}]{dai2019transformer}
Zihang Dai, Zhilin Yang, Yiming Yang, Jaime Carbonell, Quoc~V Le, and Ruslan
  Salakhutdinov. 2019.
\newblock Transformer-xl: Attentive language models beyond a fixed-length
  context.
\newblock \emph{arXiv preprint arXiv:1901.02860}.

\bibitem[{Devlin et~al.(2018)Devlin, Chang, Lee, and
  Toutanova}]{devlin2018bert}
Jacob Devlin, Ming-Wei Chang, Kenton Lee, and Kristina Toutanova. 2018.
\newblock Bert: Pre-training of deep bidirectional transformers for language
  understanding.
\newblock \emph{arXiv preprint arXiv:1810.04805}.

\bibitem[{Gomez et~al.(2017)Gomez, Ren, Urtasun, and
  Grosse}]{gomez2017reversible}
Aidan~N Gomez, Mengye Ren, Raquel Urtasun, and Roger~B Grosse. 2017.
\newblock The reversible residual network: Backpropagation without storing
  activations.
\newblock In \emph{Advances in neural information processing systems}, pages
  2214--2224.

\bibitem[{Guo et~al.(2019)Guo, Qiu, Liu, Shao, Xue, and Zhang}]{guo2019star}
Qipeng Guo, Xipeng Qiu, Pengfei Liu, Yunfan Shao, Xiangyang Xue, and Zheng
  Zhang. 2019.
\newblock Star-transformer.
\newblock \emph{arXiv preprint arXiv:1902.09113}.

\bibitem[{Huang et~al.(2018)Huang, Vaswani, Uszkoreit, Simon, Hawthorne,
  Shazeer, Dai, Hoffman, Dinculescu, and Eck}]{huang2018music}
Cheng-Zhi~Anna Huang, Ashish Vaswani, Jakob Uszkoreit, Ian Simon, Curtis
  Hawthorne, Noam Shazeer, Andrew~M Dai, Matthew~D Hoffman, Monica Dinculescu,
  and Douglas Eck. 2018.
\newblock Music transformer: Generating music with long-term structure.
\newblock In \emph{International Conference on Learning Representations}.

\bibitem[{Joshi et~al.(2019)Joshi, Levy, Weld, and Zettlemoyer}]{joshi2019bert}
Mandar Joshi, Omer Levy, Daniel~S Weld, and Luke Zettlemoyer. 2019.
\newblock Bert for coreference resolution: Baselines and analysis.
\newblock \emph{arXiv preprint arXiv:1908.09091}.

\bibitem[{Kitaev et~al.(2020)Kitaev, Kaiser, and Levskaya}]{kitaev2020reformer}
Nikita Kitaev, {\L}ukasz Kaiser, and Anselm Levskaya. 2020.
\newblock Reformer: The efficient transformer.
\newblock \emph{arXiv preprint arXiv:2001.04451}.

\bibitem[{Kwiatkowski et~al.(2019)Kwiatkowski, Palomaki, Redfield, Collins,
  Parikh, Alberti, Epstein, Polosukhin, Devlin, Lee
  et~al.}]{kwiatkowski2019natural}
Tom Kwiatkowski, Jennimaria Palomaki, Olivia Redfield, Michael Collins, Ankur
  Parikh, Chris Alberti, Danielle Epstein, Illia Polosukhin, Jacob Devlin,
  Kenton Lee, et~al. 2019.
\newblock Natural questions: a benchmark for question answering research.
\newblock \emph{Transactions of the Association for Computational Linguistics},
  7:453--466.

\bibitem[{Lai et~al.(2017)Lai, Xie, Liu, Yang, and Hovy}]{lai2017race}
Guokun Lai, Qizhe Xie, Hanxiao Liu, Yiming Yang, and Eduard Hovy. 2017.
\newblock Race: Large-scale reading comprehension dataset from examinations.
\newblock \emph{arXiv preprint arXiv:1704.04683}.

\bibitem[{Lan et~al.(2019)Lan, Chen, Goodman, Gimpel, Sharma, and
  Soricut}]{lan2019albert}
Zhenzhong Lan, Mingda Chen, Sebastian Goodman, Kevin Gimpel, Piyush Sharma, and
  Radu Soricut. 2019.
\newblock Albert: A lite bert for self-supervised learning of language
  representations.
\newblock \emph{arXiv preprint arXiv:1909.11942}.

\bibitem[{Liu et~al.(2020)Liu, Gong, Fu, Yan, Chen, Jiang, Lv, and
  Duan}]{liu2020rikinet}
Dayiheng Liu, Yeyun Gong, Jie Fu, Yu~Yan, Jiusheng Chen, Daxin Jiang, Jiancheng
  Lv, and Nan Duan. 2020.
\newblock Rikinet: Reading wikipedia pages for natural question answering.
\newblock \emph{arXiv preprint arXiv:2004.14560}.

\bibitem[{Liu et~al.(2018)Liu, Saleh, Pot, Goodrich, Sepassi, Kaiser, and
  Shazeer}]{liu2018generating}
Peter~J Liu, Mohammad Saleh, Etienne Pot, Ben Goodrich, Ryan Sepassi, Lukasz
  Kaiser, and Noam Shazeer. 2018.
\newblock Generating wikipedia by summarizing long sequences.
\newblock \emph{arXiv preprint arXiv:1801.10198}.

\bibitem[{Liu et~al.(2019)Liu, Ott, Goyal, Du, Joshi, Chen, Levy, Lewis,
  Zettlemoyer, and Stoyanov}]{liu2019roberta}
Yinhan Liu, Myle Ott, Naman Goyal, Jingfei Du, Mandar Joshi, Danqi Chen, Omer
  Levy, Mike Lewis, Luke Zettlemoyer, and Veselin Stoyanov. 2019.
\newblock Roberta: A robustly optimized bert pretraining approach.
\newblock \emph{arXiv preprint arXiv:1907.11692}.

\bibitem[{Maruf et~al.(2019)Maruf, Martins, and Haffari}]{maruf2019selective}
Sameen Maruf, Andr{\'e}~FT Martins, and Gholamreza Haffari. 2019.
\newblock Selective attention for context-aware neural machine translation.
\newblock \emph{arXiv preprint arXiv:1903.08788}.

\bibitem[{Miculicich et~al.(2018)Miculicich, Ram, Pappas, and
  Henderson}]{miculicich2018document}
Lesly Miculicich, Dhananjay Ram, Nikolaos Pappas, and James Henderson. 2018.
\newblock Document-level neural machine translation with hierarchical attention
  networks.
\newblock \emph{arXiv preprint arXiv:1809.01576}.

\bibitem[{Oord et~al.(2018)Oord, Li, and Vinyals}]{oord2018representation}
Aaron van~den Oord, Yazhe Li, and Oriol Vinyals. 2018.
\newblock Representation learning with contrastive predictive coding.
\newblock \emph{arXiv preprint arXiv:1807.03748}.

\bibitem[{Rae et~al.(2019)Rae, Potapenko, Jayakumar, and
  Lillicrap}]{rae2019compressive}
Jack~W Rae, Anna Potapenko, Siddhant~M Jayakumar, and Timothy~P Lillicrap.
  2019.
\newblock Compressive transformers for long-range sequence modelling.
\newblock \emph{arXiv preprint arXiv:1911.05507}.

\bibitem[{Raffel et~al.(2019)Raffel, Shazeer, Roberts, Lee, Narang, Matena,
  Zhou, Li, and Liu}]{raffel2019exploring}
Colin Raffel, Noam Shazeer, Adam Roberts, Katherine Lee, Sharan Narang, Michael
  Matena, Yanqi Zhou, Wei Li, and Peter~J Liu. 2019.
\newblock Exploring the limits of transfer learning with a unified text-to-text
  transformer.
\newblock \emph{arXiv preprint arXiv:1910.10683}.

\bibitem[{Rajpurkar et~al.(2016)Rajpurkar, Zhang, Lopyrev, and
  Liang}]{rajpurkar2016squad}
Pranav Rajpurkar, Jian Zhang, Konstantin Lopyrev, and Percy Liang. 2016.
\newblock Squad: 100,000+ questions for machine comprehension of text.
\newblock \emph{arXiv preprint arXiv:1606.05250}.

\bibitem[{Rothe et~al.(2020)Rothe, Narayan, and Severyn}]{rothe2020leveraging}
Sascha Rothe, Shashi Narayan, and Aliaksei Severyn. 2020.
\newblock Leveraging pre-trained checkpoints for sequence generation tasks.
\newblock \emph{Transactions of the Association for Computational Linguistics},
  8:264--280.

\bibitem[{Roy et~al.(2020)Roy, Saffar, Vaswani, and
  Grangier}]{roy2020efficient}
Aurko Roy, Mohammad Saffar, Ashish Vaswani, and David Grangier. 2020.
\newblock Efficient content-based sparse attention with routing transformers.
\newblock \emph{arXiv preprint arXiv:2003.05997}.

\bibitem[{Scarselli et~al.(2008)Scarselli, Gori, Tsoi, Hagenbuchner, and
  Monfardini}]{scarselli2008graph}
Franco Scarselli, Marco Gori, Ah~Chung Tsoi, Markus Hagenbuchner, and Gabriele
  Monfardini. 2008.
\newblock The graph neural network model.
\newblock \emph{IEEE Transactions on Neural Networks}, 20(1):61--80.

\bibitem[{Shaw et~al.(2019)Shaw, Massey, Chen, Piccinno, and
  Altun}]{shaw2019generating}
Peter Shaw, Philip Massey, Angelica Chen, Francesco Piccinno, and Yasemin
  Altun. 2019.
\newblock Generating logical forms from graph representations of text and
  entities.
\newblock \emph{arXiv preprint arXiv:1905.08407}.

\bibitem[{Shaw et~al.(2018)Shaw, Uszkoreit, and Vaswani}]{shaw2018self}
Peter Shaw, Jakob Uszkoreit, and Ashish Vaswani. 2018.
\newblock Self-attention with relative position representations.
\newblock \emph{arXiv preprint arXiv:1803.02155}.

\bibitem[{Shazeer and Stern(2018)}]{shazeer2018adafactor}
Noam Shazeer and Mitchell Stern. 2018.
\newblock Adafactor: Adaptive learning rates with sublinear memory cost.
\newblock \emph{arXiv preprint arXiv:1804.04235}.

\bibitem[{Shen et~al.(2018)Shen, Zhou, Long, Jiang, and Zhang}]{shen2018bi}
Tao Shen, Tianyi Zhou, Guodong Long, Jing Jiang, and Chengqi Zhang. 2018.
\newblock Bi-directional block self-attention for fast and memory-efficient
  sequence modeling.
\newblock \emph{arXiv preprint arXiv:1804.00857}.

\bibitem[{Sukhbaatar et~al.(2019)Sukhbaatar, Grave, Bojanowski, and
  Joulin}]{sukhbaatar2019adaptive}
Sainbayar Sukhbaatar, Edouard Grave, Piotr Bojanowski, and Armand Joulin. 2019.
\newblock Adaptive attention span in transformers.
\newblock \emph{arXiv preprint arXiv:1905.07799}.

\bibitem[{Sun et~al.(2020)Sun, Xiong, Liu, Liu, and Bao}]{sun2020joint}
Si~Sun, Chenyan Xiong, Zhenghao Liu, Zhiyuan Liu, and Jie Bao. 2020.
\newblock Joint keyphrase chunking and salience ranking with bert.
\newblock \emph{arXiv preprint arXiv:2004.13639}.

\bibitem[{Vaswani et~al.(2017)Vaswani, Shazeer, Parmar, Uszkoreit, Jones,
  Gomez, Kaiser, and Polosukhin}]{vaswani2017attention}
Ashish Vaswani, Noam Shazeer, Niki Parmar, Jakob Uszkoreit, Llion Jones,
  Aidan~N Gomez, {\L}ukasz Kaiser, and Illia Polosukhin. 2017.
\newblock Attention is all you need.
\newblock In \emph{Advances in neural information processing systems}, pages
  5998--6008.

\bibitem[{Wang et~al.(2018)Wang, Singh, Michael, Hill, Levy, and
  Bowman}]{wang2018glue}
Alex Wang, Amanpreet Singh, Julian Michael, Felix Hill, Omer Levy, and Samuel~R
  Bowman. 2018.
\newblock Glue: A multi-task benchmark and analysis platform for natural
  language understanding.
\newblock \emph{arXiv preprint arXiv:1804.07461}.

\bibitem[{Welbl et~al.(2018)Welbl, Stenetorp, and
  Riedel}]{welbl2018constructing}
Johannes Welbl, Pontus Stenetorp, and Sebastian Riedel. 2018.
\newblock Constructing datasets for multi-hop reading comprehension across
  documents.
\newblock \emph{Transactions of the Association for Computational Linguistics},
  6:287--302.

\bibitem[{Xiong et~al.(2019)Xiong, Hu, Xiong, Campos, and
  Overwijk}]{xiong2019open}
Lee Xiong, Chuan Hu, Chenyan Xiong, Daniel Campos, and Arnold Overwijk. 2019.
\newblock Open domain web keyphrase extraction beyond language modeling.
\newblock \emph{arXiv preprint arXiv:1911.02671}.

\bibitem[{Yang et~al.(2019)Yang, Dai, Yang, Carbonell, Salakhutdinov, and
  Le}]{yang2019xlnet}
Zhilin Yang, Zihang Dai, Yiming Yang, Jaime Carbonell, Russ~R Salakhutdinov,
  and Quoc~V Le. 2019.
\newblock Xlnet: Generalized autoregressive pretraining for language
  understanding.
\newblock In \emph{Advances in neural information processing systems}, pages
  5754--5764.

\bibitem[{Yang et~al.(2018)Yang, Qi, Zhang, Bengio, Cohen, Salakhutdinov, and
  Manning}]{yang2018hotpotqa}
Zhilin Yang, Peng Qi, Saizheng Zhang, Yoshua Bengio, William~W Cohen, Ruslan
  Salakhutdinov, and Christopher~D Manning. 2018.
\newblock Hotpotqa: A dataset for diverse, explainable multi-hop question
  answering.
\newblock \emph{arXiv preprint arXiv:1809.09600}.

\bibitem[{Yang et~al.(2016)Yang, Yang, Dyer, He, Smola, and
  Hovy}]{yang2016hierarchical}
Zichao Yang, Diyi Yang, Chris Dyer, Xiaodong He, Alex Smola, and Eduard Hovy.
  2016.
\newblock Hierarchical attention networks for document classification.
\newblock In \emph{Proceedings of the 2016 conference of the North American
  chapter of the association for computational linguistics: human language
  technologies}, pages 1480--1489.

\bibitem[{Ye et~al.(2019)Ye, Guo, Gan, Qiu, and Zhang}]{ye2019bp}
Zihao Ye, Qipeng Guo, Quan Gan, Xipeng Qiu, and Zheng Zhang. 2019.
\newblock Bp-transformer: Modelling long-range context via binary partitioning.
\newblock \emph{arXiv preprint arXiv:1911.04070}.

\bibitem[{You et~al.(2019)You, Li, Reddi, Hseu, Kumar, Bhojanapalli, Song,
  Demmel, Keutzer, and Hsieh}]{you2019large}
Yang You, Jing Li, Sashank Reddi, Jonathan Hseu, Sanjiv Kumar, Srinadh
  Bhojanapalli, Xiaodan Song, James Demmel, Kurt Keutzer, and Cho-Jui Hsieh.
  2019.
\newblock Large batch optimization for deep learning: Training bert in 76
  minutes.
\newblock In \emph{International Conference on Learning Representations}.

\bibitem[{Zaheer et~al.(2020)Zaheer, Guruganesh, Dubey, Ainslie, Alberti,
  Onta\~{n}\'{o}n, Pham, Ravula, Wang, Yang et~al.}]{zaheer2020big}
Manzil Zaheer, Guru Guruganesh, Avinava Dubey, Joshua Ainslie, Chris Alberti,
  Santiago Onta\~{n}\'{o}n, Philip Pham, Anirudh Ravula, Qifan Wang, Li~Yang,
  et~al. 2020.
\newblock Big bird: Transformers for longer sequences.
\newblock \emph{arXiv preprint arXiv:2007.14062}.

\bibitem[{Zhang et~al.(2019)Zhang, Wei, and Zhou}]{zhang2019hibert}
Xingxing Zhang, Furu Wei, and Ming Zhou. 2019.
\newblock Hibert: Document level pre-training of hierarchical bidirectional
  transformers for document summarization.
\newblock \emph{arXiv preprint arXiv:1905.06566}.

\end{thebibliography}

\clearpage

\appendix

\section*{Appendix A: Implementation Details}\label{app:implementation}

\subsection*{Global-Local Attention Implementation}

This appendix provides further details on the TPU/GPU-friendly implementation of global-local attention. 
Our implementation of sliding window local attention is similar to the approach in the \texttt{local\_attention\_1d} layer in Tensor2Tensor~\footnote{\url{https://arxiv.org/abs/1803.07416}}, but with the addition of flexible masking, relative position encoding, and global tokens as side keys/values. We use a simple example to describe the internal blocking logic. Let's say the input corresponds to embeddings for the following word pieces, each represented by a letter: \texttt{ABCDEFG}.

As usual, we project these embeddings into queries, keys, and values, yielding the following (for each attention head):

\begin{tabular}{ll} 
Queries: &  \texttt{A$_q$B$_q$C$_q$D$_q$E$_q$F$_q$G$_q$} \\
Keys:    &  \texttt{A$_k$B$_k$C$_k$D$_k$E$_k$F$_k$G$_k$} \\
Values:  &  \texttt{A$_v$B$_v$C$_v$D$_v$E$_v$F$_v$G$_v$} \\
\end{tabular}

Let's say we want to perform sliding window local attention with local radius $r=2$. Internally, we split the input into blocks of length $r+1$ (3 in our example) and add padding blocks to the left and right, resulting in the following five blocks for the queries (and similarly for keys and values), with \texttt{0} representing padding:

\texttt{000 A$_q$B$_q$C$_q$ D$_q$E$_q$F$_q$ G$_q$00 000}

Conceptually we'd like to compare each query with the $2r+1$ (5 in our example) surrounding keys, as follows:

\begin{tabular}{l|l} 
{\em Queries} & {\em Keys} \\ \hline
\texttt{A$_q$}  &   \texttt{00A$_k$B$_k$C$_k$} \\
\texttt{B$_q$}  &   \texttt{0A$_k$B$_k$C$_k$D$_k$} \\
\texttt{C$_q$}  &   \texttt{A$_k$B$_k$C$_k$D$_k$E$_k$} \\
\texttt{D$_q$}  &   \texttt{B$_k$C$_k$D$_k$E$_k$F$_k$} \\
\texttt{E$_q$}  &   \texttt{C$_k$D$_k$E$_k$F$_k$G$_k$} \\
\texttt{F$_q$}  &   \texttt{D$_k$E$_k$F$_k$G$_k$0} \\
\texttt{G$_q$}  &   \texttt{E$_k$F$_k$G$_k$00} \\
\end{tabular}

But materializing each window of keys would be memory-intensive. Instead, we allow each block of queries to attend to 3 blocks of keys (the same block, and the blocks immediately to the left and right), resulting in the following:

\begin{tabular}{l|l} 
{\em Queries} & {\em Keys} \\ \hline
\texttt{A$_q$B$_q$C$_q$}  & \texttt{000 A$_k$B$_k$C$_k$ D$_k$E$_k$F$_k$} \\
\texttt{D$_q$E$_q$F$_q$}  & \texttt{A$_k$B$_k$C$_k$ D$_k$E$_k$F$_k$ G$_k$00}  \\ 
\texttt{G$_q$00}          & \texttt{D$_k$E$_k$F$_k$ G$_k$00 000} \\
\end{tabular}

Now each query can potentially see a few more tokens than it's strictly allowed to by the local radius $r$.  For example, \texttt{A$_q$} takes a dot product with \texttt{D$_k$}, \texttt{E$_k$} and \texttt{F$_k$}, but this is easy to simply mask out, yielding the same sliding window local attention result. In this way, the blocking mechanism saves memory at the expense of some extra compute.

The values are also divided into the same blocks as the keys (concatenating 3 at a time), and standard scaled dot product attention is applied independently for each row in the table below, where Keys have been truncated for brevity:


 \begin{tabular}{l|l|l} 
 {\em Queries} & {\em Keys} & {\em Values} \\ \hline
 \texttt{A$_q$B$_q$C$_q$}    & ...  & \texttt{000 A$_v$B$_v$C$_v$ D$_v$E$_v$F$_v$} \\
 \texttt{D$_q$E$_q$F$_q$}    & ...   & \texttt{A$_v$B$_v$C$_v$ D$_v$E$_v$F$_v$ G$_v$00}  \\
 \texttt{G$_q$00}            & ...  & \texttt{D$_v$E$_v$F$_v$ G$_v$00 000}  \\
 \end{tabular}

\subsection*{Efficient Relative Attention Implementation}

To efficiently implement relative position encoding (a.k.a. relative attention), we take an approach similar to the optimization in Music Transformer~\cite{huang2018music} but generalized to allow arbitrary pairwise labels rather than adhering to a relative position pattern.  We briefly describe our implementation in the case of full attention (with a single sequence length $n$), but the approach naturally extends to the case of the four attention pieces used in ETC.  The original relative attention work~\cite{shaw2018self} reported $O(n^2d_z)$ memory overhead (Section \ref{subsec:longinputs}) by materializing $a_{ij}^K$ for every query-key pair while sharing $a_{ij}^K$ across all heads (or $O(hn^2d_z)$ if not sharing across heads), where $d_z$ is the dimension per head and $h$ is the number of heads.  We instead take a dot product between each query vector and all unique $a_{ij}^K$ vectors in the relative attention vocabulary.  Then we gather these scalar results for each query-key pair.  This avoids the $O(n^2d_z)$ memory overhead and allows us to use different $a_{ij}^K$ per attention head with no additional activation memory cost. Note that our relative attention vocabulary sizes are noticeably smaller than $n$, so our implementation reduces the number of dot products required for relative attention also.

\section*{Appendix B: CPC Loss in ETC}\label{app:cpc}

We adapted the original formulation of CPC for ETC by modeling it as a dual encoder problem. We have two instances of the same ETC model $g_{1}$ and $g_{2}$ (using the same weights). $g_{1}$ is the main encoder we are training, and we divide its long input into segments (e.g., sentences) and have one global token in global input for each segment. We mask some segments in the long input, and encode those segments independently using $g_{2}$ (by having as input just the tokens of that segment in the long input, and a single global token in the global input). Then, we train $g_{1}$ and $g_{2}$ so that the encodings of the global tokens corresponding to the masked segments should be as similar as possible as the encoding of the global segment token obtained via $g_{2}$. We use within-batch random negatives for this process as well, and use a Noise Contrastive Estimation (NCE) loss, in the same way as in the original CPC work~\cite{oord2018representation}.

\section*{Appendix C: Training Details}\label{app:training}

Our default pre-training procedure used the same Wikipedia and Books training corpora as BERT, but we filtered to remove those documents with fewer than 7 sentences. Models were pre-trained for 33 epochs to match the amount of pre-training of the original BERT model, which used batches of 256 sequences of 512 tokens each, and pre-trained for 1,000,000 iterations. Instead, we used batches of 512 sequences of 4096 tokens each and pre-trained for 63,000 iterations.  The ETC-large models were pre-trained for 66 epochs by using a batch size of 1024 instead. When lifting weights from RoBERTa, we found that decreasing the learning rate to $2 \times 10^{-3}$ improved model quality.

When pre-training models, we split any input documents that are longer than the long input length. For efficiency, we also concatenate as many shorter documents as will fit into the 512/4096/8192 window and mask attention to prevent them from influencing each other. This results in a roughly 3x speedup in pre-training time for 4096-token models, highlighting once more the advantage of flexible masking.

When pre-training with CPC, we randomly select 10\% of sentences to be masked for the CPC task. Subsequently, 15\% of the remaining tokens are masked for MLM.

In the models where we use both MLM and CPC, we used a 0.8 weight for MLM and a 0.2 weight for CPC to combine them into a single loss.

\subsection*{NQ}

{\bf Data Download Link:} \url{https://ai.google.com/research/NaturalQuestions/download}

{\bf Data Pre-processing:} Following Alberti's BERT implementation~\cite{alberti2019bert}, long input in NQ contains a CLS token followed by the question word pieces, then a separator followed by the long document, a final separator, and padding. Global input contains a CLS token, one special ``question'' token per token in the question, and then one special ``segment'' token per paragraph (long answer candidate) in the long input. Moreover, since the ground truth indexes in this dataset are word indexes, in order to be able to align tokens with words, sentences are first tokenized by words, and then each word is given to the BERT/RoBERTa tokenizer. 

{\bf Fine-Tuning:} After pre-training, all models were fine-tuned with a hyperparameter sweep consisting of learning rates in $\{3\times10^{-5}, 5\times10^{-5}\}$ and number of epochs in $\{3, 5\}$ ($\{2, 3\}$ for large models) with a batch size of 64 on the NQ training set using the Adam optimizer. The model is trained to predict four logits coming out of the long input tokens: long answer start, long answer end, short answer start, and short answer end. A final prediction (predicted from the long input CLS token embedding) is the answer type (null, yes, no, short, long). For NQ instances that are longer than long input size, a sliding window approach is used (with stride 128 for input lengths of 512, 2048 for input lengths of 4096, and 4096 for input lengths of 8192).

{\bf Model Selection:} we performed a single hyper-parameter sweep for each model (i.e., we tested a single random seed per parameter configuration). The best model was selected as the highest average F1 score on dev at the end of fine-tuning. Our best model (large, lifting from RoBERTa) was trained for 2 epochs with learning rate $3\times10^{-5}$.

{\bf Inference:} Final predictions are then aggregated similarly as in the work of \citet{alberti2019bert}, but with two improvements: First, instead of predicting start and end of short/long answer separately, we first select the best start position, and then select the best end location that occurs after the start position. For short answer, we also filter out all end positions further than 38 words from the start position. Second, when the logit for a yes/no answer is higher than the logits for short, long or null answer, we replace the short answer with a corresponding yes/no.

\subsection*{HotpotQA}

{\bf Data Download Link:} \url{https://hotpotqa.github.io/}

{\bf Data Pre-processing:}  Only 90,431 out of the 90,447 instances were used for training, as we model the task as extractive QA and thus filtered out the 16 instances where the answer could not be found in the contexts. Long input in HotpotQA is organized as follows: CLS token followed by question tokens followed by all the context tokens. Each context is represented as the concatenation of its title, and then all the sentences. Global input has a CLS token, and then one token per question token (as in NQ), followed by global tokens representing the contexts. For every context, in global, we have one token representing the whole context, and then one per sentence. We did not use any windowing approach for longer instances, and just fit as many tokens as possible within the 4096 long input. Global input length was set to 256. 

{\bf Fine-Tuning:} After pre-training, all base models were fine-tuned with a hyperparameter sweep consisting of learning rates in $\{3\times10^{-5}, 5\times10^{-5}\}$, number of epochs in $\{3, 5, 7, 9\}$, batch size in $\{32, 64\}$, and supporting fact threshold in $\mathit{linspace}(0, 1, 11)$ on the training set using the Adam optimizer. Large models were tested with learning rate in $\{1\times10^{-5}, 2\times10^{-5}, 3\times10^{-5}, 5\times10^{-5}, 7\times10^{-5}\}$, number of epochs in \{2, 5, 9, 13\}, and batch size in \{32, 64, 128\}. 

{\bf Model Selection:} we performed a single hyper-parameter sweep to determine the best parameter configuration for each model, and then we tried 3 different random seeds for the best configuration. The best model was selected as the one with the best joint F1 score on dev. Our best model (large, lifting from RoBERTa) was trained for 5 epochs, with a learning rate of $3\times10^{-5}$, batch size of 32 and supporting fact threshold of 0.4.

{\bf Inference:} In order to make predictions, supporting facts are predicted using a single dense layer taking the global input embeddings as input with a threshold over the output logits. Output type is predicted with a single dense layer from the global CLS token. Answer spans where predicted also with dense layers, but using the long input embeddings as inputs, using the following criteria: begin/end positions must be in sentences or titles, begin/end must be in the same sentence/title, spans must belong to a supporting fact, begin must be before end, and spans cannot exceed a maximum answer length of 30 tokens. Within spans satisfying those criteria, a single span with top ${\rm begin\_prob} * {\rm end\_prob}$ is selected.

\subsection*{WikiHop}

{\bf Data Download Link:} \url{https://qangaroo.cs.ucl.ac.uk/}

{\bf Data Pre-processing:} Global and Long input was set similarly as in HotpotQA, except that global input was set to 430, and that instead of a CLS token in global, we have one token per candidate answer (WikiHop provides a list of candidate answers, and the model needs to select among them). We used a relative position label (the same used to link sentence summary tokens with its corresponding tokens) to link candidate answers to their mentions in the text, where mentions are determined only by string matching. Also, as in HotpotQA, no sliding window was used, and instances were just cropped to a length of 4096. A MaxHeap was used to ensure that in case a context is truncated, truncation happens from the contexts with the larger number of sentences. 

{\bf Fine-Tuning:} After pre-training, all base models were fine-tuned with a hyperparameter sweep consisting of learning rates in $\{1\times10^{-5}, 2\times10^{-5}, 3\times10^{-5}, 4\times10^{-5},  5\times10^{-5}\}$, and number of epochs in $\{5, 10, 15\}$ with a batch size of 64 on the training set using the Adam optimizer. For large models, we narrowed down the hyperparameter sweep to learning rates in $\{2\times10^{-5}, 3\times10^{-5}, 4\times10^{-5},  5\times10^{-5}\}$, and number of epochs in $\{5, 10\}$. For this dataset we also experimented with the LAMB optimizer (in addition to Adam), which was used for our leaderboard submission. 

{\bf Model Selection:} we performed a single hyper-parameter sweep to determine the best parameter configuration for each model (i.e., single random seed per parameter configuration). The best model was selected as the one with the highest accuracy on dev. Our best model (large, lifting from RoBERTa) was trained for 10 epochs, with a learning rate of $5\times10^{-5}$. Finally, for the final leaderboard submission, we selected the checkpoint of the model that had the highest dev set accuracy.

{\bf Inference:} For final prediction, we used a dense layer from the global input embeddings, after that, the candidate with the highest logit is selected as the final prediction.

\subsection*{OpenKP} 

{\bf Data Download Link:} \url{https://microsoft.github.io/msmarco/}

{\bf Data Pre-processing:} Long input in OpenKP contains all the word pieces of the input. One global token per VDOM node was added to the global input (notice this is like the pre-training setup, except instead of sentences we have VDOM nodes as the higher-level units). No sliding windowing was used, and we simply truncate instances to whichever of max tokens in long input or max VDOM tokens in global ends up being more constraining. Long input length was set to 4096 and global input length to 512 by default in this dataset. After url deduplication and skipping examples without keyphrases in the truncated document, there were 133,374 valid training examples. Regarding visual features, we embed font sizes based on 24 bucket ranges, and we also construct an embedding for the cross of ``block'', ``heading'', and ``bolded'' Boolean features in the input data.  All other visual features were treated as dense features, with the floating point features clipped to reasonable ranges and re-scaled to the $[-1, 1]$ interval.  These dense features are then transformed to the same embedding space as the other embeddings, and all visual feature embeddings are added to both the relevant long and global input tokens. 

{\bf Fine-Tuning:} After pre-training, all models were fine-tuned with a hyperparameter sweep consisting of learning rates in $\{3\times10^{-5}, 5\times10^{-5}\}$ and number of epochs in $\{2, 3\}$ with a batch size of 64 on the OpenKP training set using the Adam optimizer. To generate predictions,  we first sum the embeddings (from the long input) of all the word pieces for each word to form word embeddings. Then we run convolutions with kernel size 1, 2, 3, 4, and 5 to form the respective $n$-gram embeddings. Finally, a dense linear layer is used to form logits for all the $n$-grams and concatenate them together for one combined softmax. The loss is cross entropy where the ground truth probabilities are divided equally among the keyphrase labels (up to 3). By default we used the first occurrence of each keyphrase as the label. Our improved ``max loss'' takes the max of logits across all occurrences of the same keyphrase in the text, rather than just the first occurrence. 

{\bf Model Selection:} we performed a single hyper-parameter sweep to determine the best parameter configuration for each model (i.e., single random seed per parameter configuration). The best model was selected as the one with the highest F1@3 on dev. Our best model (large, max loss, lifting from RoBERTa) was trained for 2 epochs, with a learning rate of $3\times10^{-5}$.

{\bf Inference:} During inference, we select the top 5 keyphrases ordered by logits, removing any duplicates. All keyphrases were treated as uncased for the purpose of deduplication.

\section*{Appendix D: Lifting Weights from BERT/RoBERTa}

When lifting weights from BERT or RoBERTa, the weights that can be lifted are (for every Transformer layer): feed forward layer, $W^Q$, $W^K$, $W^V$ (since BERT/RoBERTa only have one copy of such matrices, in models where we use different matrices for global and long inputs, we initialize both sets of matrices to the same BERT/RoBERTa weights), attention output projection, and layer normalization. Additionally, we can also lift the token embedding matrix. Absolute position embeddings and next sentence prediction weights from BERT are discarded. After that, weights for the layers necessary for the CPC loss, and those involved in relative position encodings are randomly initialized.

\begin{table}[tb]\centering 
\begin{tabular}{lrr} 
{\em Model} & {\em Time} & {\em Hardware}   \\ \hline
ETC (share) & 11h 13m   & 256 core TPU v3 \\
ETC         & 11h 46m   & 256 core TPU v3 \\
ETC-large   & 63h 41m   & 512 core TPU v3 \\
\end{tabular}							
\caption{Time taken for pre-training the different model types used in our experiments, together with the hardware configuration used (2 cores = 1 chip). This corresponds to 63k pre-training iterations, with batch size 512 for base models (33 epochs), and 1024 for large models (66 epochs).}
\label{tbl:times} 
\end{table}

\begin{table}[tb]\centering
\resizebox{\columnwidth}{!}{%
\begin{tabular}{lrrr} 
{\em Dataset} & {\em Epochs} & {\em Time} & {\em Hardware}   \\ \hline
NQ        & 5   &   10h 47m   & 32 core TPU v3 \\
HotpotQA  & 9   &   2h 59m   & 32 core TPU v3 \\
WikiHop   & 15  &   5h 55m   & 32 core TPU v3 \\
OpenKP    & 3   &   2h 5m & 32 core TPU v3 \\
\end{tabular}%
}
\caption{Time taken for fine-tuning the baseline ETC (base) model on different datasets, together with the hardware configuration used (2 cores = 1 chip). As we did a hyper-parameter sweep with different number of epochs, we report the time of the largest number of epochs we tried.}
\label{tbl:times-ft} 
\end{table}

\begin{figure}[t!]
	\includegraphics[width=\columnwidth]{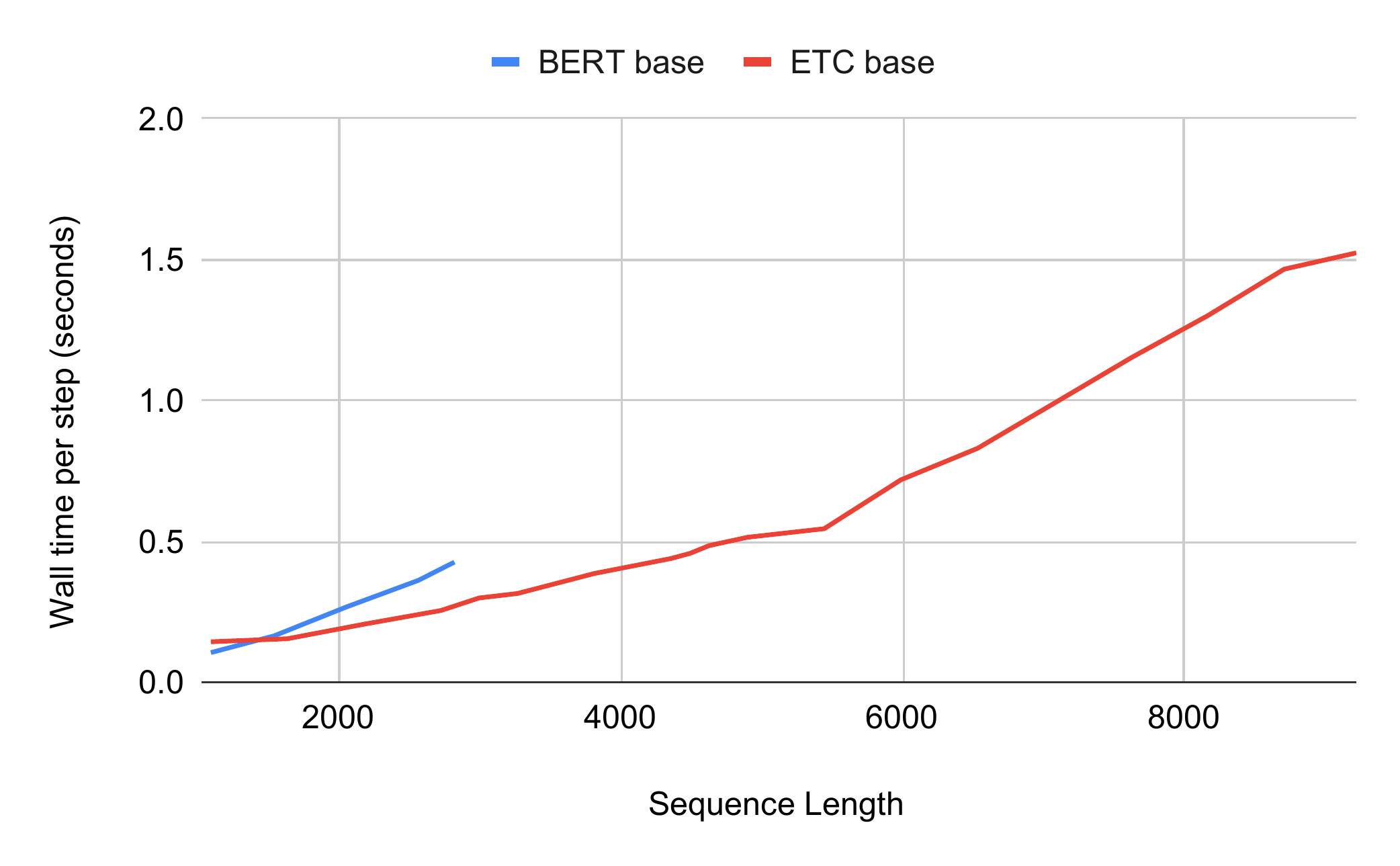}
	\centering
	\caption{Wall time per step for different input lengths for both BERT and ETC with their base configurations. For ETC, global input length was set to $1/{16}$th of the long input length until reaching a ceiling of 512 global length at 8192 long length, and Sequence Length is the sum of long and global lengths.}
	\label{fig:walltime}
\end{figure}

\begin{figure}[tb]
	\includegraphics[width=\columnwidth]{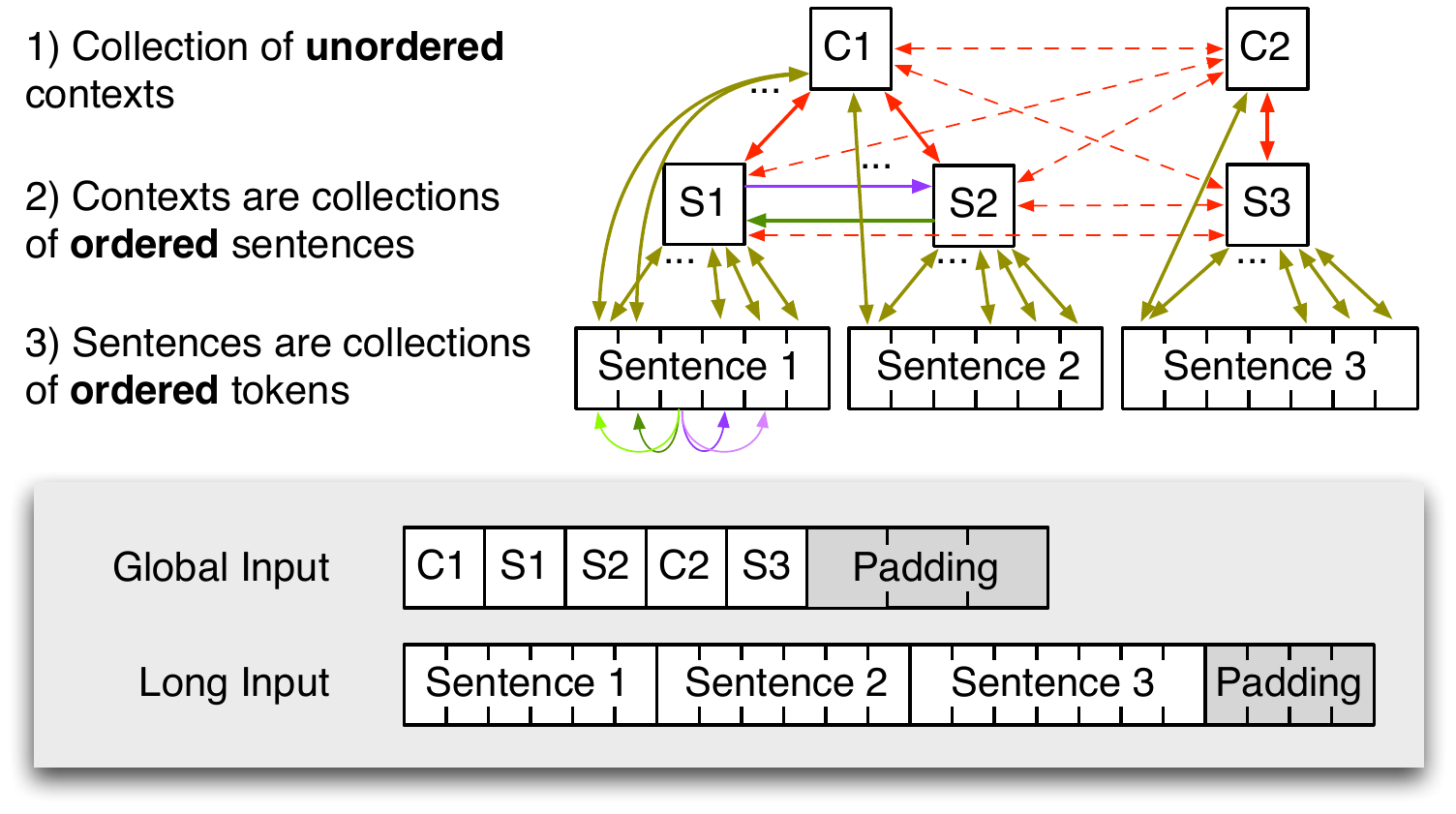}
	\centering
	\caption{Illustration of structure encoding with ETC (using the same example shown in Figure \ref{fig:etc-attention2}b).
	Top: each box represents an input token, and arrows represent attention. The different colors and dash patterns in arrows represent the different relative position labels. Bottom: illustration of where each token would appear in the input of ETC.}
	\label{fig:etc-structure}
\end{figure}

For lifting to be possible, the number of layers, hidden size, number of attention heads, and size of the feed forward intermediate layers of the BERT/RoBERTa model need to match with the ETC model.

\section*{Appendix E: Model Computational Requirements}\label{app:parameters}

{\bf Memory:} To gauge headroom for scaling input lengths beyond what we used in this paper, we ran some additional experiments on TPU v3 hardware with gradient checkpointing and removing the extra gradient moments required by optimizers like Adam and LAMB. Fixing global input length to 512 tokens, we were able to push {\em base} models to long input size of 22656, and {\em large} models to long input size of 8448 before running out of memory on a single TPU v3 core. We leave for future work experimentation with more memory-efficient optimizers like Adafactor~\cite{shazeer2018adafactor} and model-parallelism techniques in ETC.

{\bf Compute:} As stated above, the computational complexity of attention in ETC is $O(n_g (n_g + n_l) + n_l (n_g + 2r+1 ))$ and if we assume $n_g = O(2r+1)$, this results in a complexity of $O(n_g^2 + n_g n_l)$, which is linear in the size of the long input. Table \ref{tbl:times} shows pre-training times in our experiments, together with the hardware used in each experiment. Table \ref{tbl:times-ft} shows the fine-tuning times taken by the baseline ETC model on the different datasets. Notice that pre-training is the most computational intensive part, and thus, we used significantly more hardware. In order to gain further insights into the common use case of running the models using GPUs, Figure \ref{fig:walltime} shows a comparison of the wall-time used per step when using a single NVIDIA Tesla V100 GPU as the input length increases, for both BERT and ETC in their base configurations. As the plot shows ETC is initially slower, but it becomes faster than BERT for input lengths larger than about 1500. Moreover, the BERT plot ends earlier due to memory constraints. Finally, notice that the ETC wall time is not linear in this figure, as we also increased the size of the global input together with the long input.


\begin{table}[tb]\centering 
\begin{tabular}{lcccc} 
{\em Model} & {\em Parameters} \\ \hline
ETC base (shared)   &   109M \\
ETC base            &   166M \\
ETC large (RoBERTa vocab) &   558M  \\ \hline
BERT base                   &  110M  \\
BERT large (RoBERTa vocab)  &  355M  \\  
\end{tabular}							
\caption{Number of trainable parameters for the different models evaluated in this paper.}
\label{tbl:parameters} 
\end{table}


{\bf Parameters:} Finally, Table \ref{tbl:parameters} shows the total number of parameters of the ETC model for the different configurations used in our experiments. The most important consideration is that the number of trainable parameters does not depend on the input length. As a matter of fact, it only depends on: the embedding dimensionality ($d$), the number of layers ($l$), and the number of relative position labels (which depends on $k$).
The parameter count also depends on the fully connected feed forward intermediate size, but this is $4d$ by convention and for all ETC models in this paper. Our baseline model uses separate $W^Q$, $W^K$, $W^V$, and output projection matrices for global and long inputs, resulting in about 50\% more parameters than BERT. But the configuration with shared  $W^Q$, $W^K$, $W^V$, and output projection has a similar number of parameters as BERT. Parameter count for BERT base is reported from the original paper~\cite{devlin2018bert} and BERT large (RoBERTa vocab) from the original RoBERTa paper~\cite{liu2019roberta}.



\section*{Appendix F: Structured Input Example}\label{app:structure}

Figure \ref{fig:etc-attention2}b shows an illustration of a possible attention pattern for a dataset like WikiHop, where the input consists of contexts, made out of sentences. There is no order among the contexts, but there is among the sentences within a context. Figure \ref{fig:etc-structure} illustrates how this can be encoded in ETC, putting all the word piece tokens in the long input, and using the global input for special ``context'' and ``sentence'' tokens. Different relative position labels are used to indicate the different relations (token part of a sentence, sentence part of a context, order between sentences, order between tokens, etc.).

\end{document}